\documentclass[a4paper]{article}
\usepackage{interspeech2013,amssymb,amsmath,epsfig}
\ninept
\def\reg{{\rm\ooalign{\hfil
     \raise.07ex\hbox{\scriptsize R}\hfil\crcr\mathhexbox20D}}}

\title{Model-based Bayesian Reinforcement Learning for Dialogue Management}

\makeatletter
\def\name#1{\gdef\@name{#1\\}}
\makeatother
\name{}
\name{{\em Pierre Lison$^1$}}

\usepackage{wrapfig}
\usepackage[tight,footnotesize]{subfigure}
\usepackage{algorithmic}
\usepackage{algorithm}
\DeclareMathOperator*{\argmax}{arg\,max}

\address{\\}
\address{$^1$Language Technology Group, Department for Informatics, University of Oslo, Norway\\ {\small \tt plison@ifi.uio.no}} 

\begin{document}
\maketitle
\begin{abstract}
Reinforcement learning methods are increasingly used to optimise dialogue policies from experience.  Most current techniques are \textit{model-free}: they directly estimate the utility of various actions, without explicit model of the interaction dynamics.  In this paper, we investigate an alternative strategy grounded in \textit{model-based Bayesian reinforcement learning}.  Bayesian inference is used to maintain a posterior distribution over the model parameters, reflecting the model uncertainty.  This parameter distribution is gradually refined as more data is collected and simultaneously used to plan the agent's actions. \\
Within this learning framework, we carried out experiments with two alternative formalisations of the transition model, one encoded with standard multinomial distributions, and one structured with probabilistic rules. We demonstrate the potential of our approach with empirical results on a user simulator constructed from Wizard-of-Oz data in a human--robot interaction scenario.  The results illustrate in particular the benefits of capturing prior domain knowledge with high-level rules.
\end{abstract}
\noindent{\bf Index Terms}: dialogue management, reinforcement learning, Bayesian inference, probabilistic models, POMDPs

\section{Introduction}

Designing good control policies for spoken dialogue systems can be a daunting task, due both to the pervasiveness of speech recognition errors and the large number of dialogue trajectories that need to be considered.  
In order to automate part of the development cycle and make it less prone to design errors, an increasing number of approaches have come to rely on reinforcement learning (RL) techniques \cite{FramptonL09,Supelec270,InTech_RL_2008_OP,gasic2011,Cuayahuitl:2010,Henderson:2008,Thomson:2010:BUD:1772996.1773040,Young:2010}  to automatically optimise the dialogue policy. The key idea is to model dialogue management as a Markov Decision Process (MDP) or a Partially Observable Markov Decision Process (POMDP), and let the system learn by itself the best action to perform in each possible conversational situation via repeated interactions with a (real or simulated) user. Empirical studies have shown that policies optimised via RL are generally more robust, flexible and adaptive than their hand-crafted counterparts \cite{Supelec270,6407655}.  

To date, most reinforcement learning approaches to policy optimisation have adopted \textit{model-free} methods such as Monte Carlo estimation \cite{Young:2010}, Kalman Temporal Differences \cite{DBLP:journals/tslp/PietquinGCF11}, SARSA($\lambda$) \cite{Henderson:2008}, or Natural Actor Critic \cite{Jurcicek:2011}.  In model-free methods, the learner seeks to directly estimate the expected return ($Q$-value) for every state-action pairs based on the set of interactions it has gathered.  The optimal policy is then simply defined as the one that maximises this $Q$-value. 

In this paper, we explore an alternative approach, inspired by recent developments in the RL community: \textit{model-based Bayesian reinforcement learning} \cite{Ross:2011,poupart2008}.  In this framework, the learner doesn't directly estimate Q-values, but rather gradually constructs an explicit model of the domain in the form of transition, reward and observation models.  Starting with some initial priors, the learner iteratively refines the parameter estimates using standard Bayesian inference given the observed data.  These parameters are then subsequently used to plan the optimal action to perform, taking into consideration every possible source of uncertainty (state uncertainty, stochastic action effects, and model uncertainty).

In addition to providing an elegant, principled solution to the exploration-exploitation dilemma \cite{Ross:2011}, model-based Bayesian RL has the additional benefit of allowing the system designer to directly incorporate his/her prior knowledge into the domain models.  This is especially relevant for dialogue management, since many domains exhibit a rich internal structure with multiple tasks to perform, sophisticated user models, and a complex, dynamic context.  We argue in particular that models encoded via probabilistic rules can boost learning performance compared to unstructured distributions.  

The contributions of this paper are twofold.  We first demonstrate how to apply model-based Bayesian RL to learn the transition model of a dialogue domain. We also compare two modelling approaches in the context of a human--robot scenario where a Nao robot is instructed to move around and pick up objects. The empirical results show that the use of structured representations enables the learning algorithm to converge faster and with better generalisation performance.  

The paper is structured as follows.  $\S$\ref{background} reviews the key concepts of reinforcement learning.  We then describe how model-based Bayesian RL operates ($\S$\ref{mbbrl})  and detail two alternative formalisations for the domain models ($\S$\ref{modelling}). We evaluate the learning performance of the two models in $\S$\ref{evaluation}.  $\S$\ref{previouswork} compares our approach with previous work, and $\S$\ref{conclusion} concludes.

\section{Background}
\label{background}

\subsection{POMDPs}

Drawing on previous work \cite{Williams:2007,Thomson:2010:BUD:1772996.1773040,Young:2010,5946754,daubigney2012}, we  formalise dialogue management as a \textit{Partially Observable Markov Decision Process} (POMDP) $\langle S, A, O, T, Z, R \rangle$, where $S$ represents the set of possible dialogue states $s$, $A$ the set of system actions $a_m$, and $O$ the set of observations -- here, the N-best lists that can be generated by the speech recogniser.   $T$ is the transition model $P(s'|s,a_m)$ determining the probability of reaching state $s'$ after executing action $a_m$ in state $s$.  $Z$ is the probability $P(o|s)$ of observing $o$ when the current (hidden) state is $s$.  Finally, $R(s,a_m)$ is the reward function, which defines the immediate reward received after executing action $a_m$ in state $s$.

In POMDPs, the current state is not directly observable by the agent, but is inferred from the observations.  The agent knowledge at a given time is represented by the \textit{belief state} $b$, which is a probability distribution $P(s)$ over possible states.  After each system action $a_m$ and subsequent observation $o$, the belief state $b$ is updated to incorporate the new information:
\begin{equation}
b'(s)\!=\!P(s'|b, a_m,o)\!=\!\alpha P(o|s') \sum_{s} P(s'|s,a_m) b(s) \label{update1}
\end{equation}
where $\alpha$ is a normalisation constant.  

In line with other approaches \cite{Thomson:2010:BUD:1772996.1773040}, we represent the belief state as a Bayesian Network and factor the state $s$ into three distinct variables $s = \langle a_u, i_u, c \rangle$, where $a_u$ is the last user dialogue act, $i_u$ the current user intention, and $c$ the interaction context.   Assuming that the observation $o$ only depends on the last user act $a_u$, and that $a_u$ depends on both the user intention $i_u$ and the last system action $a_m$,  Eq. (\ref{update1}) is rewritten as:
\begin{align}
&b'(a_u, i_u, c) = P(a_u', i_u', c'|b, a_m,o)\\
&= \alpha P(o|a_u') P(a_u'|i_u', a_m) \sum_{i_u, c} P(i_u'|i_u,a_m, c) b(i_u, c)
\end{align}
$P(o|a_u')$ is often defined as $P(\tilde{a}_u)$, the dialogue act probability in the N-best list provided by the speech recognition and semantic parsing modules.   $P(a_u'|i_u', a_m)$ is called the \textit{user action model}, while $P(i_u'|i_u, a_m, c)$ is the \textit{user goal model}. 

\subsection{Decision-making with POMDPs}

The agent objective is to find the action $a_m$ that maximise its expected cumulative reward $Q$. Given a belief state--action sequence $[ b_0, a_0, b_1, a_1,..., b_n, a_n ]$ and a discount factor $\gamma$, the expected cumulative reward is defined as:
\begin{equation}
Q([ b_0, a_0, b_1, a_1, ... b_n, a_n ] ) = \sum_{t=0}^n \gamma^t R(b_t, a_t) 
\end{equation}
where $R(b,a) = \sum_{s \in S} R(s,a) b(s)$. Using the fixed point of Bellman's equation \cite{Bellman:1957}, the expected return for the optimal policy can be written in the following recursive form:
\begin{equation}
Q(b,a) =R(b,a) + \sum_{o \in O} P(o|b,a) \max_{a'} Q(b',a') 
\end{equation}
where $b'$ is the updated dialogue state following the execution of action $a$ and the observation of $o$, as in Eq. \ref{update1}.   For notational convenience, we used $P(o|b,a) = \sum_{s \in S} P(o|s,a) b(s)$.  

If the transition, observation and reward models are known, it is possible to apply POMDP solution techniques to extract an optimal policy $\pi: b \rightarrow a$ mapping from a belief point to the action yielding the maximum $Q$-value \cite{Pineau_2003,KurHsu08,NIPS2010_0740}.

Unfortunately, for most dialogue domains, these models are not known in advance.  It is therefore necessary to collect a large amount of interactions in order to estimate the optimal action for each given (belief) state.  This is typically done by trial-and-error, exploring the effect of all possible actions and gradually focussing the search on those yielding a high return  \cite{citeulike:112017}. Due to the number of interactions that are necessary to reach convergence, most approaches rely on \textit{user simulators} for the policy optimisation.  These user simulators are often bootstrapped from Wizard-of-Oz experiments in which the system is remotely controlled by a human expert \cite{DBLP:phd/de/Rieser2008}. 

\section{Approach}
\label{mbbrl}

Contrary to model-free methods that directly estimate the policy or $Q$-value of (belief) state--action pairs, model-based Bayesian reinforcement learning relies on explicit transition, reward and observation models.  These models are gradually estimated from the data collected by the learning agent, and are simultaneously used to plan the actions to execute.  Model estimation and decision-making are therefore intertwined.  

\subsection{Bayesian learning }

The estimation of the model parameters is done via Bayesian inference -- that is, the learning algorithm  maintains a posterior distribution over the parameters $\mathbf{\theta}$ of the POMDP models, and updates these parameters given the evidence.

We focus in this paper on the estimation of the transition model $P(s'|s, a_m)$. It should however be noted that the same approach can in principle be applied to estimate the observation and reward models \cite{Ross:2011}.  The transition model can be described as a collection of multinomials (one for each possible conditional assignment of $s$ and $a_m$). It is therefore convenient to describe their parameters with Dirichlet distributions, which are the conjugate prior of multinomials.

\begin{wrapfigure}[17]{r}{32mm} 
\vspace{-0.4cm}
\begin{flushleft}
\hspace{-2mm}\includegraphics[scale=0.15]{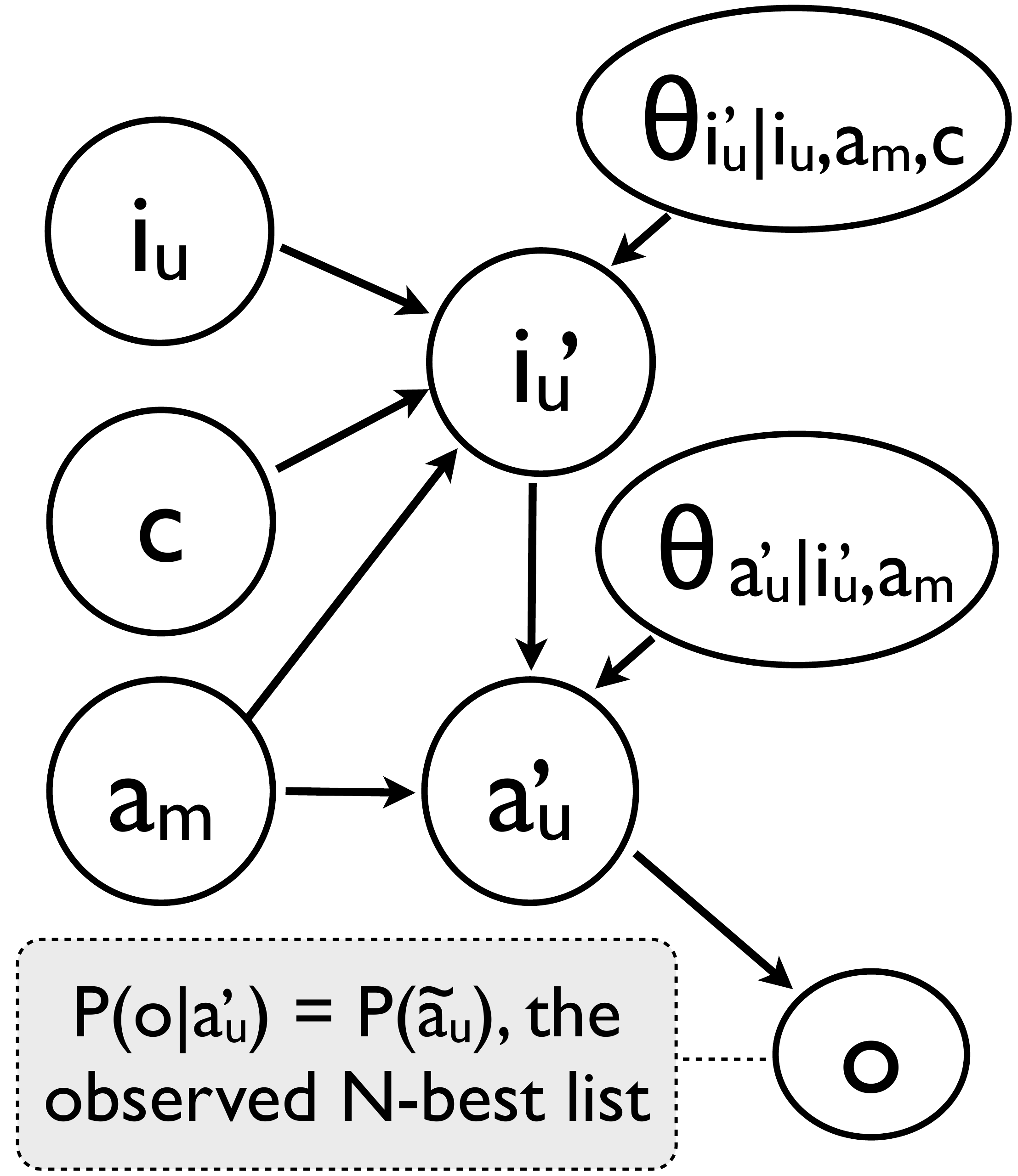}
\end{flushleft}
 \vspace{-0.5cm}
\caption{Bayesian parameter estimation of the transition model.}
\label{fig:graphmodel}
\end{wrapfigure}
Fig. \ref{fig:graphmodel} illustrates this estimation process.  The two parameters $\theta_{i_u'|i_u,am}$ and $\theta_{a_u'|i_u',a_m,c}$ respectively represent the Dirichlet distributions for the user goal and user action models.  Once a new N-best list of user dialogue acts is received, these parameters are updated using Bayes' rule, i.e. $
P(\theta|o) = \alpha P(o| \theta)$. 

The operation is repeated for every observed user act. To ensure the algorithm remains tractable, we assume conditional independence between the parameters, and we approximate the inference via importance sampling.  

\subsection{Online planning}

After updating its belief state and parameters, the agent must find the optimal action to execute, which is the one that maximises its expected cumulative reward.  This planning step is the computational bottleneck in Bayesian reinforcement learning, since the agent needs to reason not only over all the current and future states, but also over all possible transition models (parametrised by the $\mathbf{\theta}$ variables).  The high dimensionality of the task usually prevents the use of offline solution techniques.  But several approximate methods for online POMDP planning have been developed \cite{ross2008}.  In this work, we used a simple forward planning algorithm coupled with importance sampling.

\begin{algorithm}[h!]
\caption{: \small{\textsc{Q}} ($b, a, h)$}
\begin{algorithmic}[1] \vspace{1mm}
\STATE $q \leftarrow \sum_{s} b(s) R(s,a)$
\IF {$h > 1$}
\STATE $b' \leftarrow \sum_{s} P(s'|s,a) b(s)$
\STATE $v = 0$
\FOR {observation $o \in O$}
\STATE $b'' \leftarrow \sum_{s} P(o|s) b'(s)$
\STATE Estimate $Q(b'', a', h\!-\!1)$ for all actions $a'$
\STATE $v \leftarrow v + P(o|b') \max_{a'} Q(b'', a', h\!-\!1)$
\ENDFOR
\STATE $q \leftarrow q + \gamma \  v$
\ENDIF
\RETURN $q$
\end{algorithmic} 
\label{algo2}
\end{algorithm}

Algorithm \ref{algo2} shows the iterative calculation of the $Q$-value for a belief state $b$, action $a$ and planning horizon $h$.  The algorithm starts by computing the immediate reward, and then estimates the expected future reward after the execution of the action.  Line 5 loops on possible observations following the action (for efficiency reasons, only a limited number of high-probability observations are selected), and for each, the belief state is updated and its maximum expected reward is computed.  The procedure stops when the planning horizon has been reached, or the algorithm has run out of time. The planner then simply selects the action $a^* = \argmax Q(b, a)$. 

\section{Models}
\label{modelling}

We now describe two alternative modelling approaches developed for the transition model. 

\subsection{Model 1: multinomial distributions}

The first model  is constructed using standard multinomial distributions, based on the factorisation described in $\S$2.1.

Both the user action model $P(a_u'|i_u',a_m)$ and the user goal model $P(i_u'|i_u, a_m,c)$ are defined as collections of multinomials whose parameters are encoded with Dirichlet distributions.  It is possible to exploit prior domain knowledge about the relative likelihood of some event by adapting the $\alpha$ Dirichlet counts to skew the distribution in a particular direction.  For instance, we can encode the fact that the user is unlikely to change his intention after a clarification request by associating a higher $\alpha$ value to the intention $i_u'$ corresponding to the current value $i_u$ when $a_m$ is a clarification request. 

\subsection{Model 2: probabilistic rules}

The second model relies on \textit{probabilistic rules} to capture the domain structure in a compact manner and thereby reduce the number of parameters to estimate. We provide here a very brief overview of the formalism, previously presented in \cite{rulebasedmodels-sigdial2012,lison-semdial2012}.

Probabilistic rules take the form of \textit{if...then...else} control structures and map a list of conditions on input variables to specific effects on output variables.  A rule is formally expressed as an ordered list $\langle c_1, ... c_n\rangle$, where each case $c_i$ is associated with a condition $\phi_i$ and a distribution over  effects $\{(\psi_i^1, p_i^1),...,(\psi_i^k, p_i^k)\}$, where $\psi_i^j$ is an effect with associated probability $p_i^j = P(\psi_i^j | \phi_i)$. Note that $p_i^{1...m}$ must satisfy the usual probability axioms.  The rule reads as such:
\begin{align*}
& \textbf{if} \ (\phi_{1})  \ \textbf{then} \\ 
& \;\;\;\;\; \{[P(\psi_1^1) = p_1^1], \; ... \; [P(\psi_1^k) = p_1^k]\}\\
& ... \\
& \textbf{else if} \ (\phi_{n})  \ \textbf{then} \\ 
& \;\;\;\;\; \{[P(\psi_n^1) = p_n^1], \; ... \; [P(\psi_n^m) = p_n^m]\}
\end{align*}

The conditions $\phi_i$ are arbitrarily complex logical formulae grounded in the input variables.  Associated to each condition stands a list of alternative effects that define specific \textit{value assignments} for the output variables.  Each effect is assigned a probability that can be either hard-coded or correspond to a Dirichlet parameter to estimate (as in our case). 

Here is a simple example of probabilistic rule:
\begin{align*}
\textbf{Rule}: \ \ & \textbf{if} \ (a_m=\mathsf{Confirm(X)} \land i_u \neq \mathsf{X})  \ \textbf{then} \\ 
& \;\;\;\;\; \{[P(a_u' = \mathsf{Disconfirm}) = \theta_1]\}
\end{align*}
The rule specifies that, if the system requests the user to confirm that his intention is $X$, but the actual intention is different, the user will utter a $\mathsf{Disconfirm}$ action with  probability $\theta_1$ (which is presumably quite high).  Otherwise, the rule produces a void effect -- i.e. it leaves the distribution $P(a_u')$ unchanged.  

At runtime, the rules are instantiated as additional nodes in the Bayesian Network encoding the belief state.  They therefore function as high-level \textit{templates} for a plain probabilistic model. We refer once more the reader to  \cite{rulebasedmodels-sigdial2012,lison-semdial2012} for details. 

\section{Evaluation}
\label{evaluation}

We evaluated our approach within a human--robot interaction scenario.  We started by gathering empirical data for our dialogue domain using Wizard-of-Oz experiments, after which we built a user simulator on the basis of the collected data.  The learning performance of the two models was finally evaluated on the basis of this user simulator. 

\subsection{Wizard-of-Oz data collection}

\begin{wrapfigure}[13]{o}{32mm}
\vspace{-0.8cm}
\begin{center}
\includegraphics[scale=0.05]{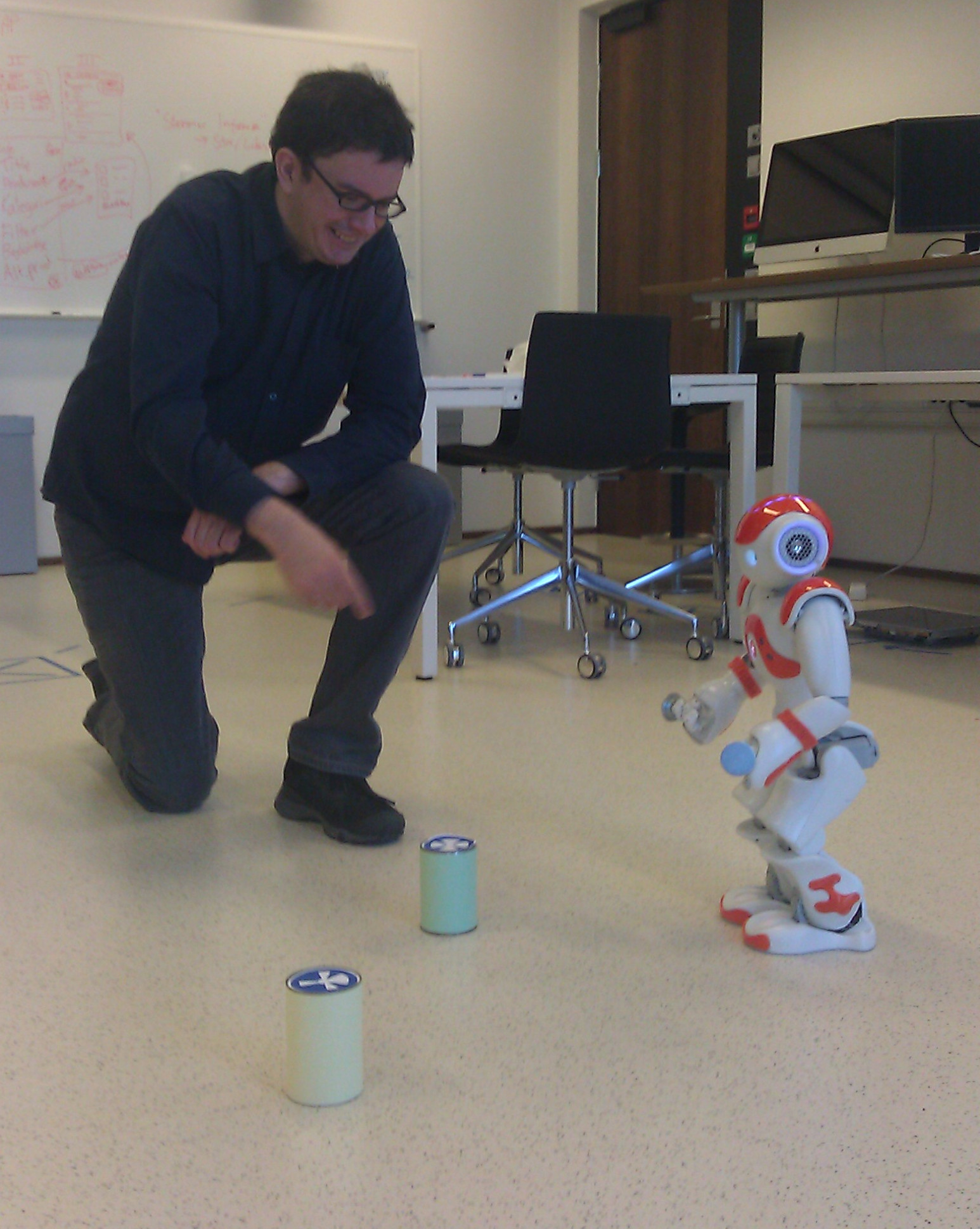}
\end{center} \vspace{-0.5cm}
\caption{User interacting with the Nao robot.}
\label{fig:nao}
\end{wrapfigure}

The dialogue domain involved a Nao robot conversing with a human user in a shared visual scene including a few graspable objects, as illustrated in Fig. \ref{fig:nao}.  The users were instructed to command the robot to walk in different directions and carry the objects from one place to another. The robot could also answer questions about his current perception (e.g. "do you see a blue cylinder?").   In total, the domain included 11 distinct user intentions, and the user inputs were classified into 16 dialogue acts. With the contextual variables, the domain included 2112 possible states. The robot could execute 37 possible actions, including both physical and conversational actions.  

8 interactions were recorded, each with a different speaker, totalling about 50 minutes.  The interactions were performed in English. After the recording, the dialogues were manually segmented and annotated with dialogue acts, system actions, user intentions, and contextual variables (e.g. perceived objects).

\subsection{User simulator}

Based on the annotated dialogues, we used MLE to derive the user goal and action models, as well as a contextual model for the robot's perception. To reproduce imperfect speech recognition, we applied a speech recogniser (Nuance Vocon) to the Wizard-of-Oz user utterances and processed the recognition results to derive a Dirichlet distribution with three dimensions respectively standing for the probability of the correct utterance, the probability of incorrect recognition, and the probability of no recognition.  The N-best lists were generated by the simulator with probabilities drawn from this distribution, estimated to $\sim\mathsf{Dirichlet}(5.4, 0.52, 1.6)$  with T. Minka's method \cite{minka2003}. 

\subsection{Experimental setup}

The simulator was coupled to the dialogue system to compare the learning performance of the two models.  The multinomial model contained 228 Dirichlet parameters.   The rule-based model contained 6 rules with 14 corresponding Dirichlet parameters.  Weakly informative priors were used for the initial parameter distributions in both models. The reward model, in Table \ref{rewards}, was identical in both cases. The planner operated with a horizon of length 2 and included an observation model introducing random noise to the user dialogue acts.

\begin{table}[h]
\begin{center}
\begin{footnotesize}
\begin{tabular}{|lll|ll|} \hline
\textit{Execution} of & correct action & +6 & wrong action & -6  \\
\textit{Answer to} & correct question & +6 & wrong question & -6  \\
\textit{Grounding} &  correct intention & +2 &  wrong intention & -6  \\ 
\textit{Ask to confirm} &  correct intention & -0.5 & wrong intention & -1.5  \\ 
& Ask to repeat & -1 & Ignore user act & -1.5 \\ \hline 
\end{tabular}\vspace{-0.4cm}
\end{footnotesize}
\end{center}  
\caption{Reward model designed for the domain.} 
\label{rewards}
\end{table}

\begin{figure}[h]
\begin{center}
\includegraphics[scale=0.28]{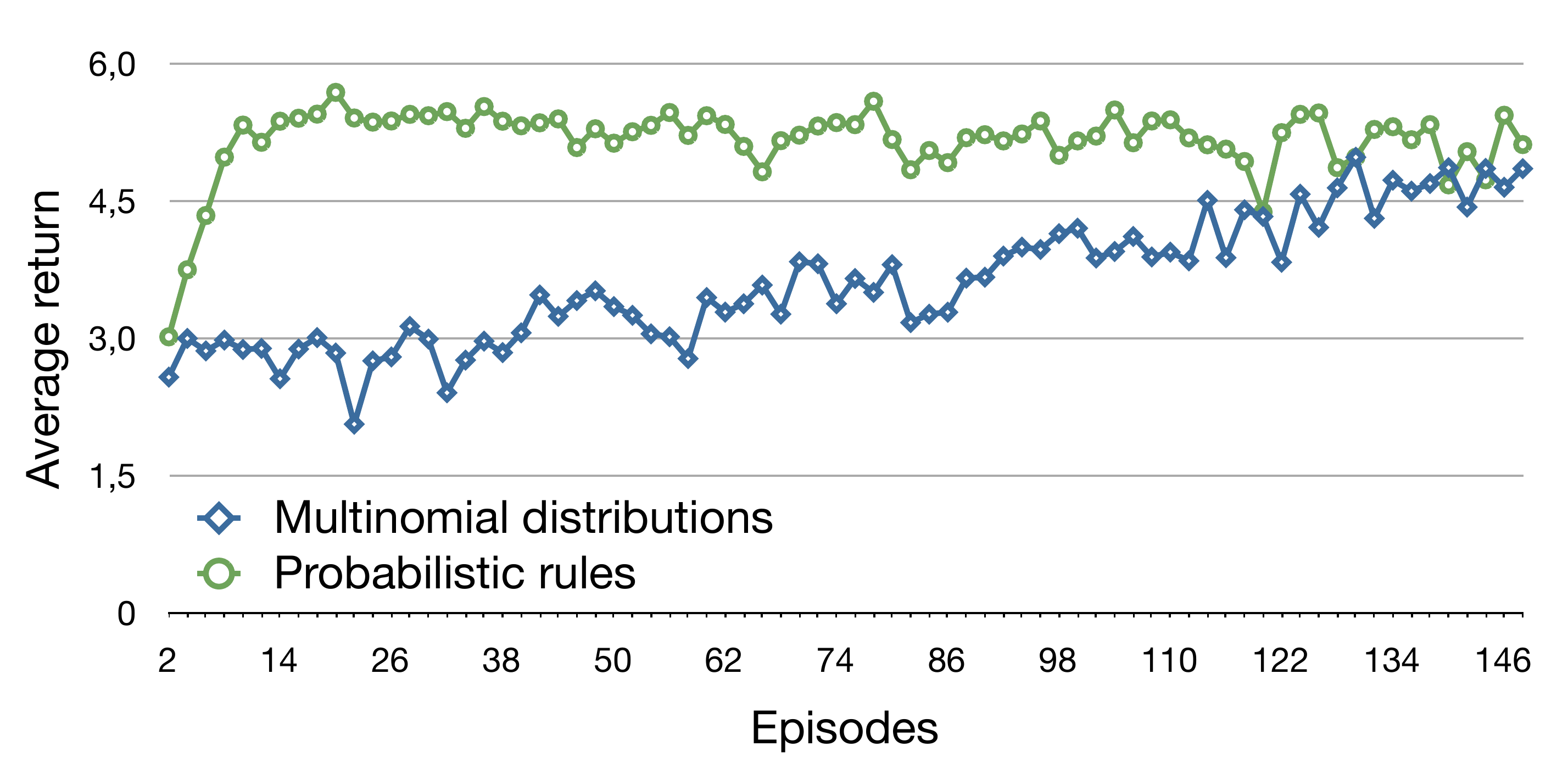}
\end{center}  \vspace{-5mm}
\caption{Average return per episode.} \vspace{-5mm}
\label{return}
\end{figure}

\begin{figure}[h] 
\begin{center}
\includegraphics[scale=0.28]{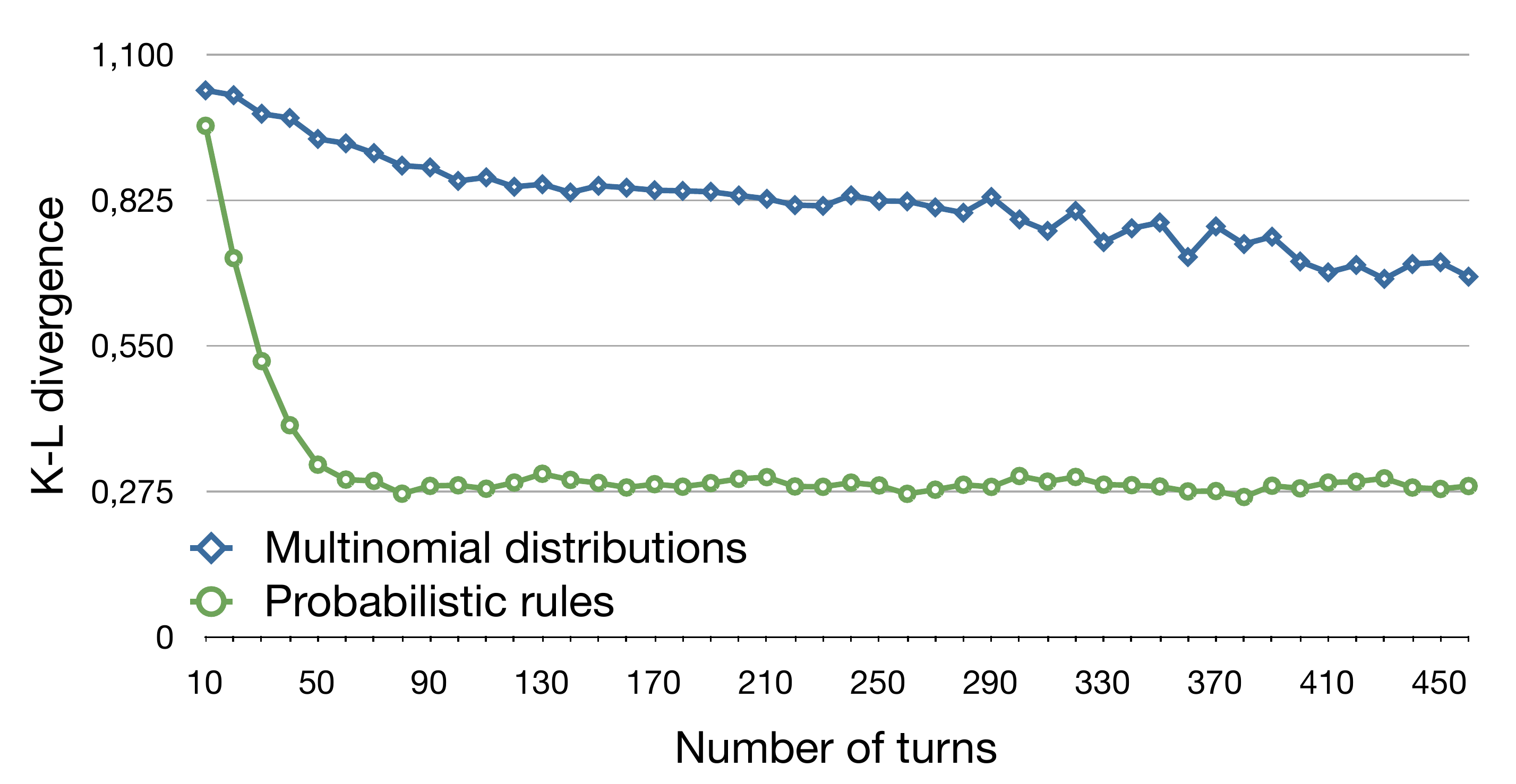}
\end{center}  \vspace{-5mm}
\caption{K-L divergence between the estimated distribution $P(a_u')$ and the actual distribution followed by the simulator.} \vspace{-2mm}
\label{divergence}
\end{figure}

The performance was first measured in terms of average return per episode, shown in Fig. \ref{return}.  To analyse the accuracy of the transition model, we also derived the Kullback-Leibler divergence \cite{KLDIVERGE} between the next user act distribution $P(a_u')$ predicted by the learned model and the actual distribution followed by the simulator at a given time\footnote{Some residual discrepancy is to be expected between these two distributions, the latter being based on the actual user intention while the former must infer it from the current belief state.} (Fig. \ref{divergence}).   The results of both figures are averaged on 100 simulations.


\subsection{Analysis of results}

The empirical results illustrate that both models are able to capture at least some of the interaction dynamics and achieve higher returns as the number of turns increases, but they do so at different learning rates.  In our view, this difference is to be explained by the higher generalisation capacity of the probabilistic rules compared to the unstructured multinomial distributions.  

It is interesting to note that most of the Dirichlet parameters associated with the probabilistic rules converge to their optimal value very rapidly, after a handful of episodes.  This is a promising result, since it implies that the proposed approach could in principle optimise dialogue policies from live interactions, without the need to rely on a user simulator, as in \cite{gasic2011}. 

\section{Related work}
\label{previouswork}

The first studies on model-based reinforcement learning for dialogue management have concentrated on learning from a fixed corpus via Dynamic Programming methods \cite{817450,Walker:2000,Singh:2000:EER:647288.723412}.  The literature also contain some recent work on Bayesian techniques. \cite{Doshi:2008:SLI:1463279.1463284} presents an interesting approach that combines Bayesian inference with active learning.  \cite{DBLP:conf/iui/AtrashP09} is another related work that utilises a sample of solved POMDP models.  Both employ offline solution techniques. To our knowledge, the only approaches based on online planning are \cite{5946754,DBLP:journals/jstsp/PngPC12}, although they focussed on the estimation of the observation model.
 
It is worth nothing that most POMDP approaches do integrate statistically estimated transition models in their belief update mechanism, but they typically do not exploit this information to optimise the dialogue policy, preferring to employ model-free methods for this purpose \cite{Young:2010,Jurcicek2012168}. 

Interesting parallels can be drawn between the structured modelling approach adopted in this paper (via the use of probability rules) and related approaches dedicated to dimensionality reduction in large state--action spaces, such as function approximation \cite{Henderson:2008}, hierarchical RL \cite{Cuayahuitl:2010}, summary POMDPs \cite{Young:2010}, state space partitioning \cite{Williams2010,Crook:2010} or relational abstractions \cite{Heriberto2011}.  These approaches are however typically engineered towards a particular type of domain (often slot-filling applications).  There has also been some work on the integration of expert knowledge using finite-state policies or ad-hoc constraints \cite{williams2008,Henderson:2008}.  In these approaches, the expert knowledge operates as an external filtering mechanism, while the probabilistic rules aim to incorporate this knowledge into the structure of the statistical model.

\section{Conclusion}
\label{conclusion}

We have presented a model-based Bayesian reinforcement learning  approach to the estimation of transition models for dialogue management. The method relies on an explicit representation of the model uncertainty via a posterior distribution over the model parameters.  Starting with an initial Dirichlet prior, this distribution is continuously refined through Bayesian inference as more data is collected by the learning agent.  An approximate online planning algorithm selects the next action to execute given the current belief state and the posterior distribution over the model parameters. 

We evaluated the approach with two alternative models, one using multinomial distributions and one based on probabilistic rules.  We conducted a learning experiment with a user simulator bootstrapped from Wizard-of-Oz data, which shows that both models improve their estimate of the domain's transition model during the interaction.  These improved estimates are also reflected in the system's action selection, which gradually yields higher returns as more episodes are completed. The probabilistic rules do however converge much faster than multinomial distributions, due to their ability to capture the domain structure in a limited number of parameters.

In our future work, we would like to directly compare our approach with model-free RL methods such as Monte-Carlo estimation or SARSA($\lambda$).  We also want to extend the framework to estimate the reward model in parallel to the state transitions. And most importantly, we plan to conduct experiments with real users to verify that the outlined approach is capable of learning dialogue policies from direct interactions.  

\newpage
\eightpt
\bibliographystyle{IEEEtran}
\bibliography{lt-biblio}

\begin{thebibliography}{10}
\providecommand{\url}[1]{#1}
\csname url@samestyle\endcsname
\providecommand{\newblock}{\relax}
\providecommand{\bibinfo}[2]{#2}
\providecommand{\BIBentrySTDinterwordspacing}{\spaceskip=0pt\relax}
\providecommand{\BIBentryALTinterwordstretchfactor}{4}
\providecommand{\BIBentryALTinterwordspacing}{\spaceskip=\fontdimen2\font plus
\BIBentryALTinterwordstretchfactor\fontdimen3\font minus
  \fontdimen4\font\relax}
\providecommand{\BIBforeignlanguage}[2]{{%
\expandafter\ifx\csname l@#1\endcsname\relax
\typeout{** WARNING: IEEEtran.bst: No hyphenation pattern has been}%
\typeout{** loaded for the language `#1'. Using the pattern for}%
\typeout{** the default language instead.}%
\else
\language=\csname l@#1\endcsname
\fi
#2}}
\providecommand{\BIBdecl}{\relax}
\BIBdecl

\bibitem{FramptonL09}
M.~Frampton and O.~Lemon, ``Recent research advances in reinforcement learning
  in spoken dialogue systems,'' \emph{Knowledge Engineering Review}, vol.~24,
  no.~4, pp. 375--408, 2009.

\bibitem{Supelec270}
O.~Lemon and O.~Pietquin, ``{Machine Learning for Spoken Dialogue Systems},''
  in \emph{{Proceedings of the 10th European Conference on Speech Communication
  and Technologies (Interspeech'07)}}, 2007, pp. 2685--2688.

\bibitem{InTech_RL_2008_OP}
O.~Pietquin, ``Optimising spoken dialogue strategies within the reinforcement
  learning paradigm,'' in \emph{Reinforcement Learning, Theory and
  Applications}.\hskip 1em plus 0.5em minus 0.4em\relax I-Tech Education and
  Publishing, 2008, pp. 239--256.

\bibitem{gasic2011}
M.~Ga\v{s}i\'{c}, F.~Jur\v{c}{\'\i}\v{c}ek, B.~Thomson, K.~Yu, and S.~Young,
  ``On-line policy optimisation of spoken dialogue systems via live interaction
  with human subjects,'' in \emph{IEEE Workshop on Automatic Speech Recognition
  and Understanding (ASRU)}, 2011, pp. 312--317.

\bibitem{Cuayahuitl:2010}
H.~Cuay\'{a}huitl, S.~Renals, O.~Lemon, and H.~Shimodaira, ``Evaluation of a
  hierarchical reinforcement learning spoken dialogue system,'' \emph{Computer
  Speech \& Language}, vol.~24, pp. 395--429, 2010.

\bibitem{Henderson:2008}
J.~Henderson, O.~Lemon, and K.~Georgila, ``Hybrid reinforcement/supervised
  learning of dialogue policies from fixed data sets,'' \emph{Computational
  Linguistics}, vol.~34, pp. 487--511, 2008.

\bibitem{Thomson:2010:BUD:1772996.1773040}
V.~Thomson and S.~Young, ``Bayesian update of dialogue state: A {POMDP}
  framework for spoken dialogue systems,'' \emph{Computer Speech \& Language},
  vol.~24, pp. 562--588, October 2010.

\bibitem{Young:2010}
S.~Young, M.~Ga\v{s}i\'{c}, S.~Keizer, F.~Mairesse, J.~Schatzmann, B.~Thomson,
  and K.~Yu, ``The hidden information state model: A practical framework for
  {POMDP}-based spoken dialogue management,'' \emph{Computer Speech \&
  Language}, vol.~24, pp. 150--174, 2010.

\bibitem{6407655}
S.~Young, M.~Ga\v{c}i\'{c}, B.~Thomson, and J.~D. Williams, ``{POMDP}-based
  statistical spoken dialog systems: A review,'' \emph{Proceedings of the
  IEEE}, vol.~PP, no.~99, pp. 1--20, 2013.

\bibitem{DBLP:journals/tslp/PietquinGCF11}
O.~Pietquin, M.~Geist, S.~Chandramohan, and H.~Frezza-Buet, ``Sample-efficient
  batch reinforcement learning for dialogue management optimization,''
  \emph{ACM Transactions on Speech \& Language Processing}, vol.~7, no.~3,
  p.~7, 2011.

\bibitem{Jurcicek:2011}
F.~Jur\v{c}\'{\i}\v{c}ek, B.~Thomson, and S.~Young, ``Natural actor and belief
  critic: Reinforcement algorithm for learning parameters of dialogue systems
  modelled as {POMDPs},'' \emph{ACM Transactions on Speech \& Language
  Processing}, vol.~7, no.~3, pp. 6:1--6:26, Jun. 2011.

\bibitem{Ross:2011}
S.~Ross, J.~Pineau, B.~Chaib-draa, and P.~Kreitmann, ``A {B}ayesian {A}pproach
  for {L}earning and {P}lanning in {P}artially {O}bservable {M}arkov {D}ecision
  {P}rocesses,'' \emph{Journal of Machine Learning Research}, vol.~12, pp.
  1729--1770, 2011.

\bibitem{poupart2008}
P.~Poupart and N.~A. Vlassis, ``Model-based bayesian reinforcement learning in
  partially observable domains,'' in \emph{International Symposium on
  Artificial Intelligence and Mathematics (ISAIM)}, 2008.

\bibitem{Williams:2007}
J.~D. Williams and S.~Young, ``Partially observable markov decision processes
  for spoken dialog systems,'' \emph{Computer Speech \& Language}, vol.~21, pp.
  393--422, 2007.

\bibitem{5946754}
S.~Png and J.~Pineau, ``Bayesian reinforcement learning for {POMDP}-based
  dialogue systems,'' in \emph{International Conference on Acoustics, Speech
  and Signal Processing (ICASSP)}, May, pp. 2156--2159.

\bibitem{daubigney2012}
L.~Daubigney, M.~Geist, and O.~Pietquin, ``Off-policy learning in large-scale
  {POMDP}-based dialogue systems,'' in \emph{International Conference on
  Acoustics, Speech and Signal Processing (ICASSP)}, 2012, pp. 4989 --4992.

\bibitem{Bellman:1957}
R.~Bellman, \emph{Dynamic programming}.\hskip 1em plus 0.5em minus 0.4em\relax
  Princeton, NY: Princeton University Press, 1957.

\bibitem{Pineau_2003}
J.~Pineau, G.~Gordon, and S.~Thrun, ``Point-based value iteration: An anytime
  algorithm for {POMDPs},'' in \emph{International Joint Conference on
  Artificial Intelligence (IJCAI)}, 2003, pp. 1025 -- 1032.

\bibitem{KurHsu08}
H.~Kurniawati, D.~Hsu, and W.~Lee, ``{SARSOP}: Efficient point-based {POMDP}
  planning by approximating optimally reachable belief spaces,'' in \emph{Proc.
  Robotics: Science and Systems}, 2008.

\bibitem{NIPS2010_0740}
D.~Silver and J.~Veness, ``Monte-carlo planning in large {POMDPs},'' in
  \emph{Advances in Neural Information Processing Systems 23}, 2010, pp.
  2164--2172.

\bibitem{citeulike:112017}
R.~S. Sutton and A.~G. Barto, \emph{{Reinforcement Learning: An
  Introduction}}.\hskip 1em plus 0.5em minus 0.4em\relax The MIT Press, 1998.

\bibitem{DBLP:phd/de/Rieser2008}
V.~Rieser, ``Bootstrapping reinforcement learning-based dialogue strategies
  from wizard-of-oz data,'' Ph.D. dissertation, Saarland University, 2008.

\bibitem{ross2008}
S.~Ross, J.~Pineau, S.~Paquet, and B.~Chaib-Draa, ``Online planning algorithms
  for {POMDPs},'' \emph{Journal of Artificial Intelligence Research}, vol.~32,
  pp. 663--704, Jul. 2008.

\bibitem{rulebasedmodels-sigdial2012}
P.~Lison, ``Probabilistic dialogue models with prior domain knowledge,'' in
  \emph{Proceedings of the SIGDIAL 2012 Conference}, 2012, pp. 179--188.

\bibitem{lison-semdial2012}
------, ``Declarative design of spoken dialogue systems with probabilistic
  rules,'' in \emph{Proceedings of the 16th Workshop on the Semantics and
  Pragmatics of Dialogue (SemDial 2012)}, 2012, pp. 97--106.

\bibitem{minka2003}
T.~Minka, ``Estimating a {D}irichlet distribution,'' \emph{Annals of Physics},
  vol. 2000, no.~8, pp. 1--13, 2003.

\bibitem{KLDIVERGE}
S.~Kullback and R.~A. Leibler, ``{On Information and Sufficiency},''
  \emph{Annals of Mathematical Statistics}, vol.~22, no.~1, pp. 79--86, 1951.

\bibitem{817450}
E.~Levin, R.~Pieraccini, and W.~Eckert, ``A stochastic model of human-machine
  interaction for learning dialog strategies,'' \emph{IEEE Transactions on
  Speech and Audio Processing}, vol.~8, no.~1, pp. 11--23, 2000.

\bibitem{Walker:2000}
M.~A. Walker, ``An application of reinforcement learning to dialogue strategy
  selection in a spoken dialogue system for email,'' \emph{Journal of
  Artificial Intelligence Research}, vol.~12, no.~1, pp. 387--416, 2000.

\bibitem{Singh:2000:EER:647288.723412}
S.~P. Singh, M.~J. Kearns, D.~J. Litman, and M.~A. Walker, ``Empirical
  evaluation of a reinforcement learning spoken dialogue system,'' in
  \emph{Proceedings of the Seventeenth National Conference on Artificial
  Intelligence}.\hskip 1em plus 0.5em minus 0.4em\relax AAAI Press, 2000, pp.
  645--651.

\bibitem{Doshi:2008:SLI:1463279.1463284}
F.~Doshi and N.~Roy, ``Spoken language interaction with model uncertainty: an
  adaptive human-robot interaction system,'' \emph{Connection Science},
  vol.~20, no.~4, pp. 299--318, Dec. 2008.

\bibitem{DBLP:conf/iui/AtrashP09}
A.~Atrash and J.~Pineau, ``A bayesian reinforcement learning approach for
  customizing human-robot interfaces,'' in \emph{Proceedings of the
  International Conference on Intelligent User Interfaces (IUI)}.\hskip 1em
  plus 0.5em minus 0.4em\relax ACM, 2009, pp. 355--360.

\bibitem{DBLP:journals/jstsp/PngPC12}
S.~Png, J.~Pineau, and B.~Chaib-draa, ``Building adaptive dialogue systems via
  bayes-adaptive {POMDPs},'' \emph{Journal of Selected Topics in Signal
  Processing}, vol.~6, no.~8, pp. 917--927, 2012.

\bibitem{Jurcicek2012168}
F.~Jur\v{c}{\'\i}\v{c}ek, B.~Thomson, and S.~Young, ``Reinforcement learning
  for parameter estimation in statistical spoken dialogue systems,''
  \emph{Computer Speech \& Language}, vol.~26, no.~3, pp. 168 -- 192, 2012.

\bibitem{Williams2010}
J.~D. Williams, ``Incremental partition recombination for efficient tracking of
  multiple dialog states,'' in \emph{Proceedings of the IEEE International
  Conference on Acoustics, Speech and Signal Processing (ICASSP)}, 2010, pp.
  5382--5385.

\bibitem{Crook:2010}
P.~A. Crook and O.~Lemon, ``Representing uncertainty about complex user goals
  in statistical dialogue systems,'' in \emph{Proceedings of the 11th SIGDIAL
  meeting on Discourse and Dialogue}, 2010, pp. 209--212.

\bibitem{Heriberto2011}
H.~Cuay\'{a}huitl, ``{L}earning {D}ialogue {A}gents with {B}ayesian
  {R}elational {S}tate {R}epresentations.'' in \emph{Proceedings of the IJCAI
  Workshop on Knowledge and Reasoning in Practical Dialogue Systems
  (IJCAI-KRPDS)}, Barcelona, Spain, 2011.

\bibitem{williams2008}
J.~D. Williams, ``{The best of both worlds: Unifying conventional dialog
  systems and {POMDPs}},'' in \emph{International Conference on Speech and
  Language Processing ({ICSLP 2008})}, Brisbane, Australia, 2008.

\end{thebibliography}

\end{document}